\documentclass{article}
\usepackage{ijcai21}

\usepackage{times}

\usepackage{soul}
\usepackage{url}
\usepackage[hidelinks]{hyperref}
\usepackage[small]{caption}
\usepackage{amsmath}
\usepackage{booktabs}
\usepackage{booktabs} 

\usepackage[english]{babel}
\usepackage{moresize}
\usepackage{amsmath}

\usepackage{algorithmic} 
\usepackage{balance}
\usepackage{comment}
\usepackage{paralist}
\usepackage{multirow}
\usepackage{arydshln}
\usepackage{filecontents}

\usepackage[english]{babel}
\usepackage[latin1]{inputenc}
\usepackage{graphicx}
\usepackage{amssymb}
\usepackage{url}
\usepackage{subfigure}
\usepackage{amsmath}
\usepackage{enumitem}
\usepackage[linesnumbered,algoruled,boxed,lined]{algorithm2e}
\usepackage{adjustbox}
\usepackage{amssymb}
\usepackage{times}
\usepackage{hyperref}

\usepackage{pgfplots}
\usetikzlibrary{pgfplots.dateplot}

\usepackage{filecontents}
\definecolor{tblue}{RGB}{31,119,180}
\definecolor{torange}{RGB}{255,127,14}
\definecolor{tgreen}{RGB}{44,160,44}
\definecolor{tred}{RGB}{214,39,40}
\definecolor{tpurple}{RGB}{148,103,189}

\usepackage{filecontents}

\newcommand{\hide}[1]{} 

\newcommand{\ie}{\textit{i}.\textit{e}.}
\newcommand{\eg}{\textit{e}.\textit{g}.} 
\newcommand{\wrt}{\textit{w}.\textit{r}.\textit{t}}

\title{Spatial-Temporal Sequential Hypergraph Network for Crime Prediction \\ with Dynamic Multiplex Relation Learning}

\author{Lianghao Xia$^1$\thanks{Both authors contribute equally to this work}, Chao Huang$^{2*}$, Yong Xu$^{1,3,4}$\thanks{Corresponding author: Yong Xu}, Peng Dai$^5$, Liefeng Bo$^5$, \\\Large{\bf Xiyue Zhang$^1$, Tianyi Chen$^1$}\\
\affiliations
$^{1}$South China University of Technology, China\\
$^{2}$University of Hong Kong, Hong Kong\\
$^{3}$Communication and Computer Network Laboratory of Guangdong, China\\
$^{4}$Peng Cheng Laboratory, China\\
$^{5}$JD Finance America Corporation, USA\\
\{cslianghao.xia,zhang.xiyue,csttychen\}@mail.scut.edu.cn, yxu@scut.edu.cn \\ chaohuang75@gmail.com,
\{peng.dai,liefeng.bo\}@jd.com
}

\def\model{ST-SHN}
\def\full{Spatial-Temporal Sequential Hypergraph Network}

\begin{document}

\maketitle

\begin{abstract}
Crime prediction is crucial for public safety and resource optimization, yet is very challenging due to two aspects: i) the dynamics of criminal patterns across time and space, crime events are distributed unevenly on both spatial and temporal domains; ii) time-evolving dependencies between different types of crimes (\eg, Theft, Robbery, Assault, Damage) which reveal fine-grained semantics of crimes. To tackle these challenges, we propose \underline{S}patial-\underline{T}emporal \underline{S}equential \underline{H}ypergraph \underline{N}etwork (\model) to collectively encode complex crime spatial-temporal patterns as well as the underlying category-wise crime semantic relationships. In specific, to handle spatial-temporal dynamics under the long-range and global context, we design a graph-structured message passing architecture with the integration of the hypergraph learning paradigm. To capture category-wise crime heterogeneous relations in a dynamic environment, we introduce a multi-channel routing mechanism to learn the time-evolving structural dependency across crime types. We conduct extensive experiments on two real-world datasets, showing that our proposed \model\ framework can significantly improve the prediction performance as compared to various state-of-the-art baselines. The source code is available at https://github.com/akaxlh/ST-SHN.
\end{abstract}

\section{Introduction}
\label{sec:intro}

Criminal activities (\eg, Burglary, Robbery and Assault) have become a major social problem due to their adverse effect on public safety and economic development~\cite{wortley2016environmental}. For example, accurate crime prediction results could facilitate the decision making process of governments and prevent crimes happening for public safety and emergency management~\cite{zhao2017modeling}. In recent years, the availability of crime data has enabled the development of data-driven methods for forecasting the occurrences of crimes~\cite{wang2016crime,2019mist}.

One solution in predicting crimes is to uncover effective features for each category of crimes based on hand-engineering domain-specific features (\eg, regional transportation information~\cite{wei2019presslight}, meteorological conditions~\cite{zheng2017contextual}). However, without domain-specific expert knowledge (\eg, criminology and anthropology), we may not have sufficient and accurate external data sources to model their correlations with different types of crimes (\eg, burglary, felony assault) precisely~\cite{regionswww17}. Therefore, a general crime data learning framework is a necessity to reduce the effort of hand-crafted feature engineering for the crime prediction task.


Among various spatial-temporal prediction methods, deep learning-based methods stand out owing to the feature representation effectiveness of neural network architecture. There exist many recently proposed forecasting frameworks focusing on modeling the time-evolving regularities over the temporal dimension and the underlying region-wise geographical dependencies over the spatial dimension, such as attentional neural methods~\cite{yao2019revisiting,huang2018deepcrime}, convolution-based learning approach~\cite{zhang2017deep}, spatial relation encoder with graph neural networks~\cite{zheng2020gman,traffic2021aaai}.

Despite their effectiveness, we argue that current spatial-temporal prediction models fall short in addressing unique challenges that are specifically for multi-dimensional crime data. Specifically, due to the crime data heterogeneity, there exist explicit and implicit dependencies among different categories of crimes. The intrinsic design of current methods, making them incapable of capturing cross-type crime influences in a fully dynamic scenario, with the integration of both spatial and temporal patterns.


\noindent \textbf{Contribution}. In light of the challenges, we propose to study the crime prediction with the goal of effectively encoding crime-type dependent representations, capturing spatial-temporal dynamics with the awareness of inherent cross-type crime influences. In this work, we present the \full\ (\model) crime prediction architecture. Technically, we develop a multi-channel routing mechanism to model the cross-type crime influences at a fine-grained level of semantics under the graph neural network framework. Additionally, the developed hierarchical spatial dependency encoder via the hypergraph learning paradigm, endows the \model\ framework with the capability of exploring region-wise relations from locally to globally. To handle the time-wise crime patterns across different types, a graph temporal shift mechanism is developed to inject the time-varying spatial-temporal crime patterns into the representation process. \model\ reveals multi-view crime patterns for better model capacity and interpretability. In a nutshell, the dynamic multi-relational modeling of crimes allows us to encode relation dependencies and semantics corresponding to time, location and crime type dimensions. We conduct extensive experiments on two datasets, to demonstrate the advantages of our model for the effectiveness of crime prediction, modeling of latent cross-type crime influence, and interpretability of multi-view relations.


\section{Preliminaries}
\label{sec:model}

We begin with some important notations and then introduce the formalized crime prediction problem. Considering a typical spatial-temporal forecasting scenario with $R$ urban regions, $T$ historical time slots (\eg, days), and $C$ crime categories (\eg~theft, assault, criminal damage), we define the crime tensor $\textbf{X}\in\mathbb{R}^{R\times T\times C}$ corresponding to the spatial, temporal and categorical dimension, respectively, in which each entry $\textbf{X}_{r,c}^t$ represents the number of $c$-th category of crime occurrences reported from region $r$ at the $t$-th time slot.


\noindent \textbf{Problem Statement}. With the aforementioned definitions, the crime prediction task can be formalized as follows: learning a prediction function that takes the historical crime tensor $\textbf{X}$ as the input, and yields predictions $\textbf{X}^{T+1}\in\mathbb{R}^{R\times C}$ on future crime occurrences. In this work, we tackle the tasks of both the binary crime occurrence prediction and forecasting the quantitative number of crime cases.


\section{Methodology}
\label{sec:solution}



\subsection{Spatial Dependency Encoder}
\model\ proposes to comprehensively capture the complex spatial dependencies across regions from different category views. We first generate the region graph $G=(\mathcal{\boldmath{V}},\mathcal{\boldmath{E}})$, where $\mathcal{\boldmath{V}}$ and $\mathcal{\boldmath{E}}$ represents the set of regions ($[r_1,...,r_R]$) and their spatially adjacent relations. Here, we use $S\times S$ grid scale to define geographical neighbors between region $r_i$ and $r_j$.


\subsubsection{Type-aware Crime Embedding Layer}
With the consideration of cross-type crime influences in a dynamic enviroment, we describe a region $r$ with multiple time-aware embeddings $\textbf{E}_{r,c}^t \in \mathbb{R}^d$ corresponding to the $c$-th type of crime and $t$-th time slot. In particular, we first normalize the crime vector $\textbf{X}_r^t$ in tensor $\textbf{X}$ using the mean $\mathbf{\mu}$ and std $\mathbf{\sigma}\in\mathbb{R}^C$ value over all $R$ regions and $T$ time slots, \ie, $\bar{\textbf{X}}_r^t=(\textbf{X}_r^t - \mathbf{\mu}) / \mathbf{\sigma}$. Formally, the time- and type-aware region embedding $\textbf{E}_{r,c}^t$ is generated as: $\textbf{E}_{r,c}^t=\bar{\textbf{X}}_{r,c}^t\cdot \textbf{e}_c$, where $\textbf{e}_c\in\mathbb{R}^d$ denotes the global embedding vector under the crime type of $c$ over all time periods, and $d$ is the hidden state size. Hence, we associate each region $r$ with $C$ different type-specific embeddings at each time slot in our dynamic multiplex spatial-temporal relation learning.


\subsubsection{Spatial Message Propagation Paradigm}
Based on the generated region embedding $\textbf{E}_{r,c}^t$ and spatial region graph $G$, we generalize our spatial relation encoder with multiplex message propagation paradigm over $G$:
\begin{align}
    m_{i\leftarrow j}^t = \bar{\textbf{A}}_{i,j} \cdot \text{Propagate}(\{\textbf{E}_{i,c}^t, \textbf{E}_{j,c}^t: c=1,...,C\})
\end{align}
\noindent where $m_{i\leftarrow j}^t$ represents the message passed from region $r_j$ to $r_i$ at $t$-th time slot. The region-wise geographical relations are reflected by the spatial adjacent matrix $\textbf{A}\in\mathbb{R}^{R\times R}$, in which $\textbf{A}_{i, j}=1$ if edge $(r_i, r_j)\in\mathcal{E}$. To debias the influence of node degree in information propagation, we further incorporate the normalization factor as $\bar{\textbf{A}}=\textbf{D}^{-\frac{1}{2}}\textbf{A}\textbf{D}^{-\frac{1}{2}}$, where $\textbf{D}$ denotes the diagonal degree matrix. Additionally, the embedding propagation function Propagate($\cdot$) is determined by both the source and target region, with the awareness of latent influence across different crime types (\eg, Burglary, Robbery).


\subsubsection{Multi-Channel Routing Mechanism} To comprehensively characterize the type-aware crime influences across different regions, we develop a routing mechanism under multiple channels as our information encoding function Propagate($\cdot$), which is formally presented as follows:
\begin{align}
    m_{i\leftarrow j}^{t,c} =\text{MC-Rout}(\textbf{E}_{i,c}^t, \textbf{E}_j^t) = \mathop{\Bigm|\Bigm|}\limits_{h=1}^H\sum_{c'=1}^C\alpha_{c,c'}^{h}\cdot\textbf{V}^h\textbf{E}_{j,c'}^t
\end{align}
\noindent In our routing mechanism, we design a multiplex mutual attention network with the learnable weight $\alpha_{c,c'}^h$. In specific, $\alpha_{c,c'}^h$ captures the explicit relevance between the $c$-th and $c'$-th crime types under the $h$-th representation subspace. $\textbf{V}^h\in\mathbb{R}^{d/H\times d}$ represents the $c$-channel transformation. $\mathop{||}$ denotes the concatenation operation. Inspired by the architecture of Transformer network~\cite{vaswani2017attention}, we propose to respectively map the target region $r_i$ and source region $r_j$ into query and key vector projection:
\begin{align}
    \bar{\alpha}_{c,c'}^h=\frac{(\textbf{Q}^h\textbf{E}_{i,c}^t)^\top(\textbf{K}^h\textbf{E}_{j,c'}^t)}{\sqrt{d/H}};~\alpha_{c,c'}^h=\frac{\text{exp}(\bar{\alpha}_{c,c'}^h)}{\sum_{c'}\text{exp}(\bar{\alpha}_{c,c'}^h)}
\end{align}
where $\textbf{Q}^h, \textbf{K}^h\in\mathbb{R}^{d/H\times d}$ are the $h$-channel transformation to acquire the query and key information.


\subsubsection{Iterative Aggregation for High-order Connectivity}
With the generated propagated message $m_{i\leftarrow j}^t\in\mathbb{R}^{C\times d}$, \model\ conducts aggregation to smooth the neighboring nodes and refine new region representation with the summation as: $\sum_{j=1}^R m_{i\leftarrow j}^t$. To extract the long-range spatial connectivity between regions, we further stack multiple spatial embedding propagation and aggregation blocks. We denote the initial region embeddings as $\textbf{E}^{(0)}$ of size $R\times T\times C\times d$, then the high-order message aggregation can be formalized as:
\begin{align}
    \textbf{E}^{(l+1)}_i=\sigma(\sum_{j=1}^R\bar{\textbf{A}}_{i,j}\cdot\text{Propagate}(\textbf{E}_i^{(l)}, \textbf{E}_j^{(l)}))
\end{align}
where $\sigma(\cdot)$ denotes the ReLU activation function with the incorporation of non-linear feature interactions. $\textbf{E}_i^{(L)}$ is obtained with $L$ iterations and is able to receive crime information from regions of $L$-hop distance.

\begin{figure}
	\centering
	\includegraphics[width=\columnwidth]{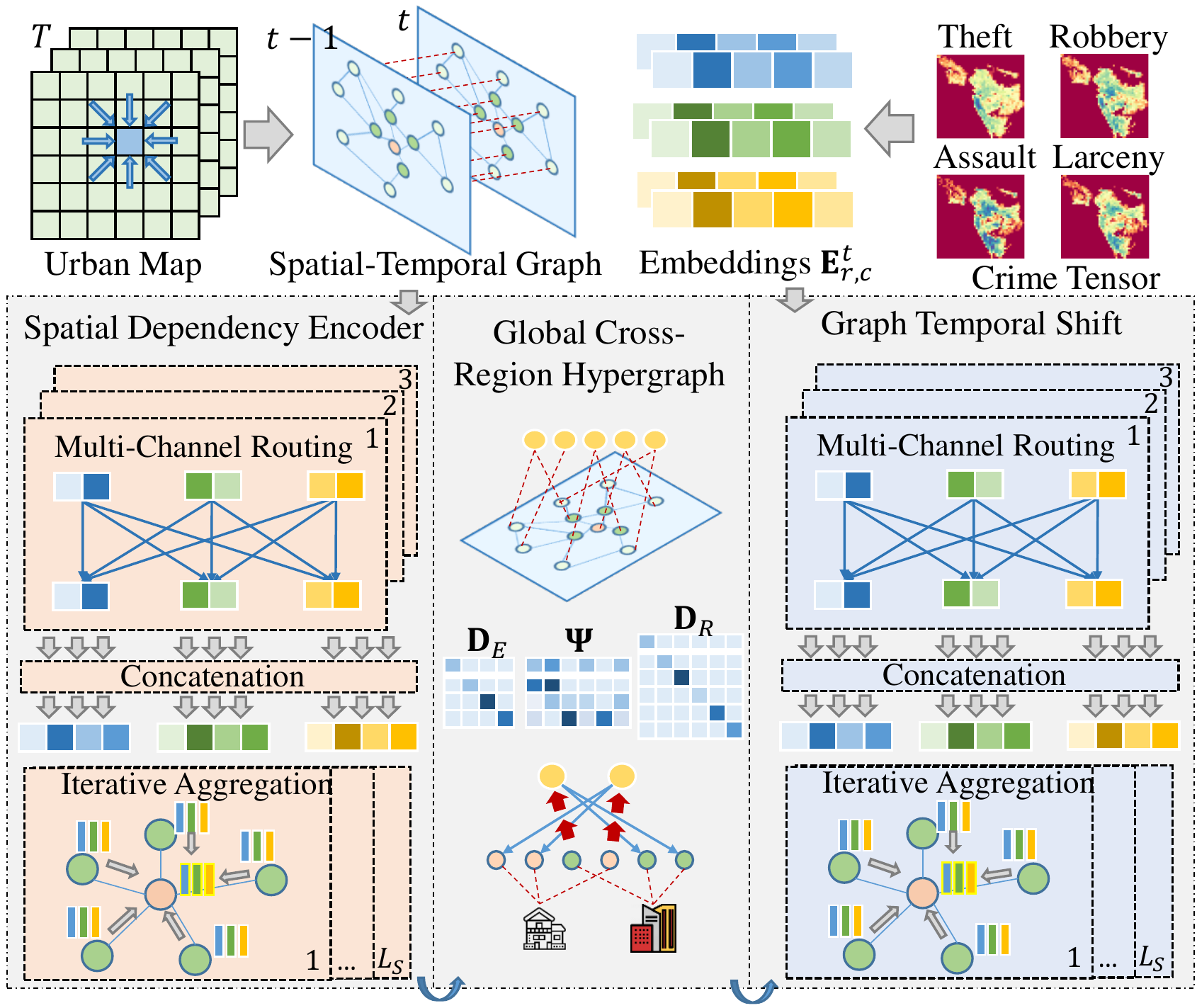}
	\caption{The model architecture of \model\ framework.}
	\label{fig:framework}
\end{figure}

\subsection{Cross-Region Hypergraph Relation Learning}
In addition to modeling spatial dependency between regions with location-aware information, we further enhance the cross-region relation learning without the limitation of adjacent connections by devising a hypergraph neural framework. With the incorporation of hypergraph neural network, we endow \model\ with the expressive power of global relation representation. Specifically, we first define a trainable adjacent matrix $\mathbf{\Psi}\in\mathbb{R}^{E\times R}$, where $E$ is a hyperparameter denoting the number of hyperedges. \model\ performs the embedding propagation along $\mathbf{\Psi}$ between $R$ regions and $E$ hyperedges. During this process, hyperedges serve as the intermediate information hubs which correlate not only locally connected but also far away regions in the entire urban space. Formally, with adjacency $\mathbf{\Psi}$ and regional embeddings $\tilde{\textbf{E}}$, the hypergraph message passing can be defined as follows:
\begin{align}
    \tilde{\textbf{E}}^{(l+1)}=\sigma(\textbf{D}_R^{-1/2} \mathbf{\Psi}^\top \textbf{D}_E^{-1/2} \sigma(\textbf{D}_E^{-1/2} \mathbf{\Psi} \textbf{D}_R^{-1/2} \tilde{\textbf{E}}^{(l)}))
\end{align}
\noindent where $\sigma(\cdot)$ denotes the activation function. $\textbf{D}_R\in\mathbb{R}^{R\times R}$, $\textbf{D}_E\in\mathbb{R}^{E\times E}$ denotes the diagonal degree matrices.

\subsection{Graph Temporal Shift Mechanism}
We further propose a graph temporal shift mechanism to encode i) intra-region type-aware crime temporal dependencies; ii) inter-region type-aware crime influences across different time slots. Towards this end, we construct a spatial-temporal crime graph between $R$ regions with the new adjacent matrix $\mathbf{\Gamma}\in\mathbb{R}^{R\times R}$, with the former dimension referring to $R$ regions at the $t$-th time slot, and the latter one corresponding to $R$ regions at the $(t+1)$-th time slot. In particular, to reflect the intra-region type-aware crime dependencies over the temporal dimension, we set $\mathbf{\Gamma}_{i,i}=1$ for region $r_i$. To enable the joint learning of inter-region and inter-type crime influence in a dynamic scenario, we set $\mathbf{\Gamma}_{i,j}=1$ between spatial adjacent region $r_i$ and $r_j$ in our graph temporal shift mechanism.



With the spatial-temporal adjacent matrix $\mathbf{\Gamma}$ and two multi-typed embedding tensors of adjacent time slots $\bar{\textbf{E}}^{(0),t}$, $\bar{\textbf{E}}^{(0),t+1}$ with the size $R\times C\times d$, \model\ recursively conducts the embedding recalibration and aggregation in a similar manner as our spatial relation encoder, as follows:
\begin{align}
    \bar{\textbf{E}}_i^{(l+1),t+1}=\sigma(\sum_{j=1}^R\bar{\mathbf{\Gamma}}_{i,j}\cdot\text{Propagate}_\text{T}(\bar{\textbf{E}}_i^{(l),t+1},\bar{\textbf{E}}_j^{(l),t}))
\end{align}
\noindent where $\text{Propagate}_\text{T}$ represents the temporal message passing paradigm which is built upon the designed multi-channel routing mechanism. $\bar{\mathbf{\Gamma}}=\textbf{D}^{-\frac{1}{2}}\mathbf{\Gamma}\textbf{D}^{-\frac{1}{2}}$ is the normalized adjacent matrix where $\textbf{D}$ is the corresponding degree matrix.


\subsection{Model Prediction and Optimization}


We integrate order-specific embeddings ($\textbf{E}^{(0)}, ..., \textbf{E}^{(L)}$) with summation. At last, \model\ sums up the embeddings along the time dimension to generate the final embeddings $\mathbf{\Lambda}\in\mathbb{R}^{R\times C\times d}$ for making predictions of different types of crimes as: $\hat{\textbf{X}}_{r,c}^{T+1}=\phi(\textbf{w}_c^\top\mathbf{\Lambda}_{r,c})$, where $\phi(\cdot)$ is Sigmoid for classification (crime occurrence) and $\phi(x)=x$ for regression (quantitative number of crime cases). For the classification and the regression tasks, \model\ is optimized by minimizing the following two loss functions, respectively:
\begin{align}
    \mathcal{L}_{c} &= -\sum_t \delta(\textbf{X}^t)  \log\hat{\textbf{X}}^t +  \bar{\delta}(\textbf{X}^t)  \log(1 - \hat{\textbf{X}}^t) + \lambda \|\mathbf{\Theta}\|_2^2\nonumber\\
    \mathcal{L}_{r}&=\sum_t\|\textbf{X}^t-\hat{\textbf{X}}^t\|_2^2 + \lambda\cdot\|\mathbf{\Theta}\|_2^2
\end{align}
where $t$ iterates the time slots, and $\delta(\cdot), \bar{\delta}(\cdot)$ denotes the element-wise positive and negative indicator functions, respectively. The later terms are the regularization terms and $\lambda$ denotes the decay weight.



\subsubsection{Model Complexity Analysis}
The spatial and the temporal pattern encoding share similar complexity: $O((|\textbf{A}|L_S+|\mathbf{\Gamma}|L_T)(C+d)TCd)$, where $|\cdot|$ denotes the number of edges in the graph, and $L_S, L_T$ denotes the number of stacked graph layers for spatial and temporal modeling, respectively. The hypergraph model costs $O(REL_HTCd)$ complexity, which is relatively minor in practice. Therefore, our model achieves the comparable time complexity as compared to other graph attentive frameworks.

\section{Evaluation}
\label{sec:eval}


This section aims to answer the following research questions:
\begin{itemize}[leftmargin=*]

\item \textbf{RQ1}: How does our developed \emph{\model} framework perform as compared to various state-of-the-art methods?

\item \textbf{RQ2}: How do different components (\eg, spatial context encoder, hypergraph relation learning, temporal shift mechanism) affect the results of \emph{\model}?


\item \textbf{RQ3}: What is the influence of parameters in \emph{\model}? 

\item \textbf{RQ4}: Can \emph{\model} provide the model interpretability \wrt\ spatial, temporal and semantic dimensions?

\end{itemize}

\begin{table}
\centering
\footnotesize

\begin{tabular}{p{1.6 cm} | p{0.5cm} | p{0.5cm} | p{0.5cm} | p{0.5cm} | p{0.5cm} | p{0.5cm} | p{0.6cm} | p{0.4cm} }
\hline
\textbf{Data} & \multicolumn{4}{c|}{\textbf{NYC-Crimes}} & \multicolumn{4}{c}{\textbf{Chicago-Crimes}} \\
\hline
Time Span & \multicolumn{4}{c|}{Jan, 2014 to Dec, 2015} & \multicolumn{4}{c}{Jan, 2016 to Dec, 2017} \\
\hline
Category & \multicolumn{2}{c|}{Burglary} & \multicolumn{2}{c|}{Robbery} & \multicolumn{2}{c|}{Theft} & \multicolumn{2}{c}{Battery} \\
\hline
Instance \# & \multicolumn{2}{c|}{31,799} & \multicolumn{2}{c|}{33,453} & \multicolumn{2}{c|}{124,630} & \multicolumn{2}{c}{99,389}\\
\hline
Category & \multicolumn{2}{c|}{Assault} & \multicolumn{2}{c|}{Larceny} & \multicolumn{2}{c|}{Damage} & \multicolumn{2}{c}{Assault}  \\
\hline
Instance \# & \multicolumn{2}{c|}{40,429} & \multicolumn{2}{c|}{85,899} & \multicolumn{2}{c|}{59,886} & \multicolumn{2}{c}{37,972} \\
\hline
\end{tabular}
\caption{Statistics of Experimented Datasets.}
\label{tab:data}
\end{table}


\subsection{Experimental Setup}

\subsubsection{Datasets}
We evaluate our model on two real-world datasets collected from the crime reporting platforms on New York City and Chicago. Each record is formatted as (crime category, timestamp and geographical coordinates). Descriptive statistics for those datasets are presented in Table~\ref{tab:data}. Each dataset contains different categories of crimes, \eg, Burglary and Robbery in NYC crime data; Theft and Damage in Chicago crime data. The reported prediction performance is averaged over 90 consecutive days during the test period of time. The grid-based mapping strategy is applied to partition NYC and Chicago into disjoint geographical regions with $3km\times 3km$ spatial unit. We use one day as the time interval to map the crime records into series. We generate the training and test set with the ratio of 7:1 and use the crimes of the last month in the training set as the validation data.

\subsubsection{Evaluation Metrics}
We adopt two types of metrics for performance evaluation. Specifically, to validate the prediction results of crime occurrences, \emph{Macro-F1} and \emph{F1-score}~\cite{huang2018deepcrime} are used for evaluating the cross-type overall performance and type-specific forecasting accuracy, respectively. To evaluate the performance in predicting quantitative number of crimes, we utilize the \emph{Mean Absolute Error (MAE)} and \emph{Mean Absolute Percentage Error (MAPE)}~\cite{geng2019spatiotemporal} as metrics.

\subsubsection{Alternative Baselines}
We compare our model with various state-of-the-art baselines which can be summarized into 5 groups. 1) conventional time series analysis techniques (ARIMA, SVM); 2) convolution-based learning model (ST-ResNet); 3) attentional spatial data prediction methods (STDN, DeepCrime, STtrans); 4) spatial-temporal prediction with graph neural networks (DCRNN, ST-GCN, GMAN); 5) deep hybrid spatial-temporal predictive solutions (UrbanFM, ST-MetaNet).

\begin{itemize}[leftmargin=*]

\item \textbf{ARIMA}~\cite{icdm12}: it is a time series analysis method which models the temporal structures with kernel-based variable regression on future values.

\item \textbf{SVM}~\cite{chang2011libsvm}: it is another time series prediction technique which transforms data into feature space. 

\item \textbf{ST-ResNet}~\cite{zhang2017deep}: it enhances the convolution neural network-based traffic prediction with the incorporation of residual network for model training efficiency.

\item \textbf{STDN}~\cite{yao2019revisiting}: it models the spatial similarity and long-term periodic temporal pattern with the designed flow gating mechanism and attention mechanism.

\item \textbf{DeepCrime}~\cite{huang2018deepcrime}: it is an attentive recurrent neural model to encode the periodicity of crime data.

\item \textbf{STtrans}~\cite{2020hierarchically}: it leverages the self-attention-based transformer network to as the information encoder for predicting crime events.

\item \textbf{DCRNN}~\cite{li2017diffusion}: it designs bidirectional random walks to capture spatial correlations and sampling-based encoder-decoder for temporal pattern modeling.

\item \textbf{ST-GCN}~\cite{yubingspatio}: it proposes to use graph convolutional layers on the graph-structured time series data to model the corresponding spatial and temporal similarities.

\item \textbf{GMAN}~\cite{zheng2020gman}: GMAN is built upon the graph-based attention network for aggregating information from both spatial and temporal dimensions.

\item \textbf{UrbanFM}~\cite{liang2019urbanfm}: it designs the convolutional feature extractor to consider local region-wise dependencies and a diffusion network to fuse meteorological factors.

\item \textbf{ST-MetaNet}~\cite{pan2019urban}: this is a meta-learning prediction approach which employs the meta knowledge from geo-graph attributes for spatial correlation modeling. 

\end{itemize}

\begin{table*}[t]
	\centering
    \footnotesize
	\setlength{\tabcolsep}{1mm}
	\begin{tabular}{|c|c|c|c|c|c|c|c|c|c|c|c|c|}
		\hline
		\multirow{3}{*}{Model} & \multicolumn{6}{c|}{New York City} & \multicolumn{6}{c|}{Chicago}\\
		\cline{2-13}
		& \multicolumn{4}{c|}{Crime Categories} & \multicolumn{2}{c|}{Overall} & \multicolumn{4}{c|}{Crime Categories} & \multicolumn{2}{c|}{Overall}\\
		\cline{2-13}
		& Burglary & Larceny & Robbery & Assault & Micro-F1 & Macro-F1 & Theft & Battery & Assault & Damage & Micro-F1 & Macro-F1\\
		\hline
		\hline
		ARIMA & 0.4054 & 0.4878 & 0.5323 & 0.4109 & 0.4591  & 0.4629 & 0.4054 & 0.4878 & 0.5323 & 0.4109 & 0.4591 & 0.4629\\
		\hline
		SVM & 0.4490 & 0.5649 & 0.5526 & 0.4262 & 0.4982 & 0.5049 & 0.6535 & 0.6434 & 0.5074 & 0.6314 & 0.6089 & 0.6068\\
		\hline
		ST-ResNet & 0.4868 & 0.6019 & 0.5638 & 0.5321 & 0.5461 & 0.5497 & 0.5551 & 0.6760 & 0.5944 & 0.6817 & 0.6268 & 0.6339\\
        \hline
        DCRNN & 0.5327 & 0.6242 & 0.6102 & 0.5187 & 0.5715 & 0.5773 & 0.6834 & 0.6469 & 0.5978 & 0.6788 & 0.6517 & 0.6507\\
        \hline
        STGCN & 0.5467 & 0.6182 & 0.6141 & 0.5100 & 0.5722 & 0.5771 & 0.6857 & 0.6894 & 0.6294 & 0.7123 & 0.6752 & 0.6755\\
        \hline
        STtrans & 0.5482 & 0.6067 & 0.6230 & 0.5289 & 0.5767 & 0.5792 & 0.7069 & 0.7106 & 0.6353 & 0.5475 & 0.6501 & 0.6498\\
        \hline
        DeepCrime & 0.4681 & 0.6470 & 0.6362 & 0.5357 & 0.5717 & 0.5820 & 0.6240 & 0.7050 & 0.6409 & 0.7312 & 0.6653 & 0.6717\\
        \hline
        STDN & 0.5255 & 0.5424 & 0.5876 & 0.4711 & 0.5316 & 0.5355 & 0.6290 & 0.6622 & 0.5864 & 0.6342 & 0.6279 & 0.6261\\
        \hline
        UrbanFM & 0.5877 & 0.4658 & 0.6385 & 0.5603 & 0.5631 & 0.5695 & 0.6885 & 0.6694 & 0.5819 & 0.6459 & 0.6464 & 0.6420\\
        \hline
        ST-MetaNet & 0.5375 & 0.5528 & 0.5354 & 0.4948 & 0.5301 & 0.5316 & 0.6983 & 0.6191 & 0.6558 & 0.7261 & 0.6748 & 0.6765\\
        \hline
        GMAN & 0.5536 & 0.5621 & 0.6256 & 0.4653 & 0.5517 & 0.5570 & 0.6162 & 0.7077 & 0.6407 & 0.7248 & 0.6723 & 0.6759\\
        \hline
        \emph{\model} & \textbf{0.6173} & \textbf{0.6503} & \textbf{0.6721} & \textbf{0.5954} & \textbf{0.6341} & \textbf{0.6348} & \textbf{0.7154} & \textbf{0.7378} & \textbf{0.6743} & \textbf{0.7293} & \textbf{0.7142} & \textbf{0.7128}\\
        \hline
	\end{tabular}
    \caption{Crime occurrence prediction performance on NYC and CHI dataset in terms of \textit{F1-score} and \textit{Macro-F1}}
	\label{tab:c_res}
\end{table*}

\subsubsection{Parameter Settings}
We implement \emph{\model} using TensorFlow and adopt Adam as the optimizer. We present the default hyperparameter settings of \emph{\model} as follows: the hidden state size $d$ is set as 16. The adjacent spatial relation scale is configured with the $3\times 3$ grid unit. In addition, the depth of our spatial path aggregation layers is set as 2. We integrate long-range temporal dependencies by stack 7 temporal shift layers. The number of latent channels is set as 4 in our multi-channel dynamic routing mechanism. We consider the number hyperedges as 128. The learning rate is initialized as 0.001 with 0.96 decay rate. We search the weight decay factor of regularization term in the range of \{0, 0.0001, 0.001, 0.01\}.


\subsection{Performance Comparison (RQ1)}
The empirical results of all methods for predicting crime occurrences (measured by F1-score and Macro-F1) and quantitative number of crime instances (measured by MAE and MAPE) are reported in Table~\ref{tab:c_res} and Table~\ref{tab:r_res}, respectively. We can find that \emph{\model} consistently outperforms all baselines on two datasets in terms of all measures. We attribute these improvements to: i) by uncovering type-aware crime influence between regions, \emph{\model} could preserve the holistic semantics of people crime behaviors; ii) benefiting from our hypergraph relational learning, \emph{\model} is able to characterize the region-wise dependencies under global urban space, which results in more powerful representations of regions; iii) with the design of graph temporal shift mechanism, \emph{\model} could better inject the dynamism of both intra- and inter-region dependencies across various crime categories underlying the spatial-temporal crime graph.

Compared with graph neural network-based spatial-temporal prediction methods (ST-GCN, GMAN, DCRNN), we can see that although they utilize location-based region graph to guide the learning of region embeddings, these approaches fail to model crime dependencies based on multiple crime categories, results in lower performance than \emph{\model}. Besides, we can observe that although DeepCrime and STtrans utilize attention networks to model category-aware crime patterns, \emph{\model} still achieves better results than them. The reason lies in that they fall short in exploring the time-evolving implicit influence among different types of crime patterns. The obvious performance gap between \emph{\model} and methods: ST-ResNet and ST-MetaNet shed light on the limitation of modeling stationary spatial dependencies.

\begin{table}[t]
	\centering
	\footnotesize

	\begin{tabular}{|c|c|c|c|c|}
		\hline
		\multirow{2}{*}{Model} & \multicolumn{2}{c|}{New York City} & \multicolumn{2}{c|}{Chicago}\\
		\cline{2-5}
		& MAE & MAPE & MAE & MAPE \\
		\hline
		\hline
		ARIMA & 0.8786 & 0.5509 & 0.8595 & 0.5056\\
		\hline
		SVM & 1.0862 & 0.6831 & 0.9899 & 0.5862\\
		\hline
		ST-ResNet & 0.9964 & 0.5799 & 0.9091 & 0.4500\\
        \hline
        DCRNN & 0.6535 & 0.4514 & 0.6807 & 0.4711\\
        \hline
        STGCN & 0.6203 & 0.4131 & 0.6822 & 0.4624\\
        \hline
        STtrans & 0.6307 & 0.4038 & 0.6890 & 0.4559\\
        \hline
        DeepCrime & 0.6231 & 0.4149& 0.6772 & 0.4551\\
        \hline
        STDN & 0.8977 & 0.4714 & 0.9530 & 0.5046\\
        \hline
        UrbanFM & 0.8095 & 0.5742 & 0.8865 & 0.6125\\
        \hline
        ST-MetaNet & 0.6243& 0.4183 & 0.7581 & 0.4795\\
        \hline
        GMAN & 0.6606 & 0.4691 & 0.7592 & 0.5438\\
        \hline
        \emph{\model} & \textbf{0.5833} & \textbf{0.3572} & \textbf{0.6425} & \textbf{0.3961}\\

        \hline
	\end{tabular}
    \caption{Forecasting results of quantitative number of crimes.}
	\label{tab:r_res}
\end{table}

\subsection{Model Ablation Study (RQ2)}
\begin{figure}[t]
	\centering
	\subfigure[][NYC F1]{
		\centering
		\includegraphics[width=0.3\columnwidth]{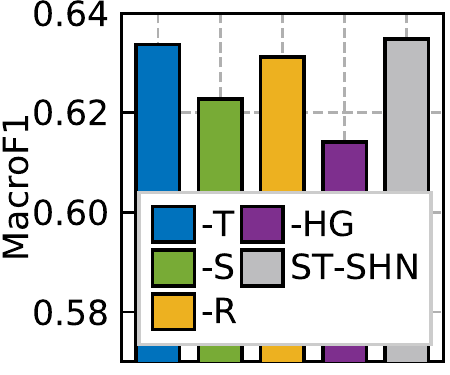}
		\label{fig:ab_nyc_micro}
	}
	\subfigure[][NYC MAE]{
		\centering
		\includegraphics[width=0.3\columnwidth]{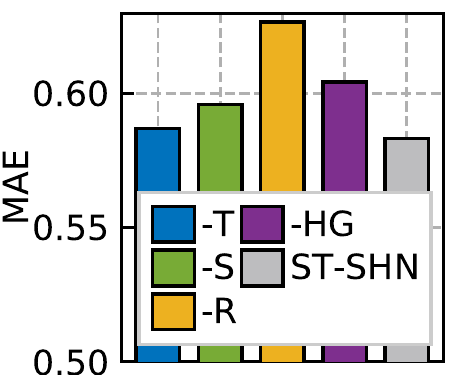}
		\label{fig:ab_nyc_mae}
	}
	\subfigure[][NYC MAPE]{
		\centering
		\includegraphics[width=0.3\columnwidth]{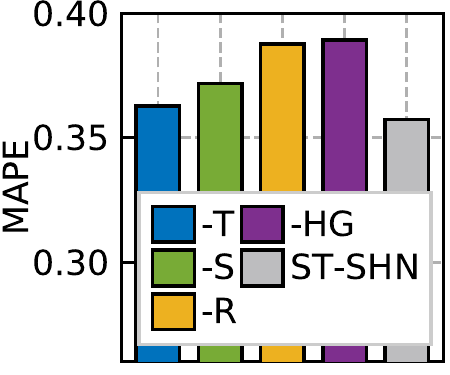}
		\label{fig:ab_nyc_mape}
	}
	\caption{Ablation studies of sub-modules in \model.}
	\label{fig:ablation}
	\vspace{-0.1in}
\end{figure}

We consider four contrast models to evaluate the effectiveness of sub-modules in \emph{\model}: 1) ``-S'': \emph{\model} without the spatial relation encoder with the location-based contextual signals; 2) ``-HG'': \emph{\model} without the hypergraph learning component to capture global region-wise relation; 3) ``-T'': \emph{\model} without the graph temporal shift mechanism; 4) ``-R'': \emph{\model} without the multi-channel routing mechanism. Figure~\ref{fig:ablation} shows the comparison results. It is lean that the full version of \emph{\model} achieves the best performance compared to other variants. Therefore, leaving cross-type crime influences across either the spatial or temporal dimension unexplored, will downgrade the prediction performance.

\subsection{Hyperparameter Study (RQ3)}

\noindent \textbf{Dimensionality of Hidden State}. The model performance is evaluated through varying the hidden dimensionality $d$ from 4 to 32. We can notice that the best performance is achieved with the channel embedding size of 16. When we conduct feature representation with latent dimensionality $>16$, the performance degrades due to the overfitting issue. \\

\noindent \textbf{Depth of Spatial-Temporal Graph Neural Model}. We search the number of graph layers for spatial, temporal dimension in the range of [0,8] and [0,14]. For spatial dimension, \emph{\model} with 2 layers outperforms the others. For temporal dimension, the best performance is achieved with the propagation path of length 8. Clearly, increasing the model depth is capable of endowing our predictive model with better representation ability. Stacking more embedding propagation layers may involve noise in refining learned representations.

\begin{figure}[t]
    \centering
    \begin{adjustbox}{max width=1.0\linewidth}
    \begin{tikzpicture}
\begin{axis}[
    xlabel={Hidden States \# $d$},
    xmin=2,xmax=34,
    ymin=0.68,ymax=0.72,
    legend columns=1,
    legend cell align=right,
    grid=both,
    every axis plot/.append style={ultra thick},
    every tick label/.append style={scale=2.4},
    label style={scale=2.6},
    legend style={
        nodes={scale=1.5, transform shape},
        legend image post style={scale=1.5},
        },
    legend style={at={(1,0)},anchor=south east},
    every axis plot post/.append style={
        every mark/.append style={scale=3.2}
    }
]
\addplot[color={rgb:blue,4;green,2;yellow,1}, mark=o, dashed, mark options={solid}]
table[x=para, y=micro] {latdim.txt};
\addplot[color={rgb:red,4;green,1;yellow,2}, mark=+, dashed, mark options={solid}]
table[x=para, y=macro] {latdim.txt};
\legend{\huge Micro-F1, \huge Macro-F1};

\end{axis}
\end{tikzpicture}

\begin{tikzpicture}
\begin{axis}[
    xlabel={Spatial Range $L_S$},
    xmin=-0.5,xmax=8.5,
    ymin=0.67,ymax=0.72,
    legend columns=1,
    legend cell align=right,
    grid=both,
    every axis plot/.append style={ultra thick},
    every tick label/.append style={scale=2.4},
    label style={scale=2.6},
    legend style={
        nodes={scale=1.5, transform shape},
        legend image post style={scale=1.5},
        },
    legend style={at={(1,0)},anchor=south east},
    every axis plot post/.append style={
        every mark/.append style={scale=3}
    }
]
\addplot[color={rgb:blue,4;green,2;yellow,1}, mark=o, dashed, mark options={solid}]
table[x=para, y=micro] {spacialRange.txt};
\addplot[color={rgb:red,4;green,1;yellow,2}, mark=+, dashed, mark options={solid}]
table[x=para, y=macro] {spacialRange.txt};
\legend{\huge Micro-F1, \huge Macro-F1};

\end{axis}
\end{tikzpicture}

\begin{tikzpicture}
\begin{axis}[
    xlabel={Temporal Range $L_T$},
    xmin=-0.5,xmax=14.5,
    ymin=0.68,ymax=0.72,
    legend columns=1,
    legend cell align=right,
    grid=both,
    every axis plot/.append style={ultra thick},
    every tick label/.append style={scale=2.4},
    label style={scale=2.6},
    legend style={
        nodes={scale=1.5, transform shape},
        legend image post style={scale=1.5},
        },
    legend style={at={(1,0)},anchor=south east},
    every axis plot post/.append style={
        every mark/.append style={scale=3}
    }
]
\addplot[color={rgb:blue,4;green,2;yellow,1}, mark=o, dashed, mark options={solid}]
table[x=para, y=micro] {temporalGnnRange.txt};
\addplot[color={rgb:red,4;green,1;yellow,2}, mark=+, dashed, mark options={solid}]
table[x=para, y=macro] {temporalGnnRange.txt};
\legend{\huge Micro-F1, \huge Macro-F1};

\end{axis}
\end{tikzpicture}
    \end{adjustbox}
    \caption{Hyperparameter study of \emph{\model}}
    \label{fig:hyperparam}
\end{figure}
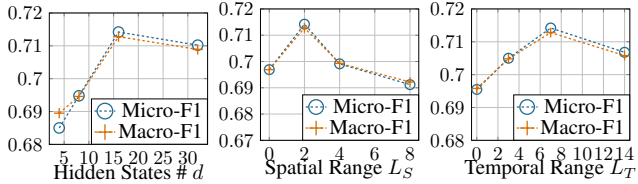

\subsection{Explainability of \emph{\model} (RQ4)}
\begin{figure}[t]
	\centering
	\includegraphics[width=0.45\textwidth]{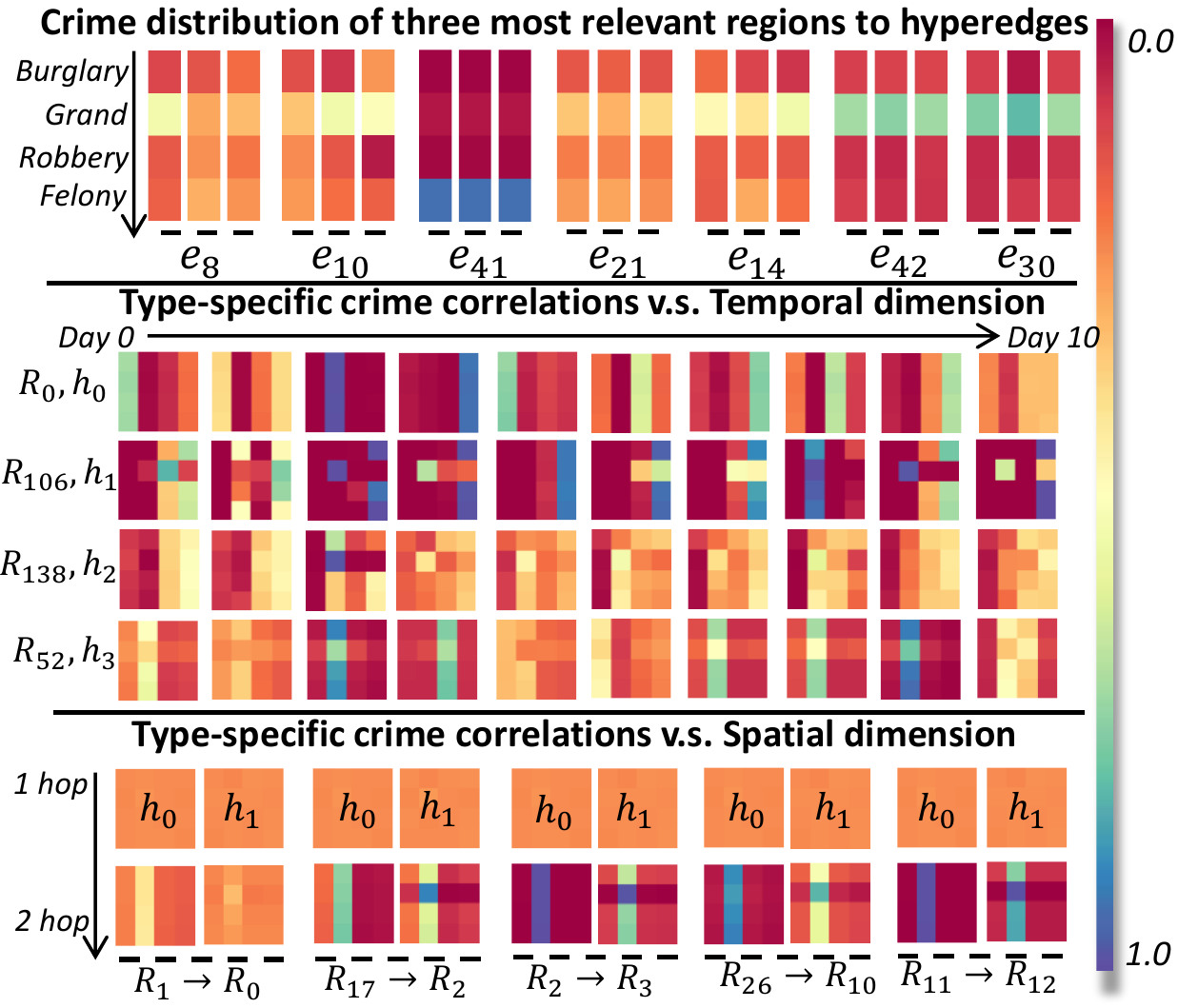}
	\caption{Visualization for learned relevance weights from the hypergraph learning, temporal and spatial relation learning. Bars represent the distribution of different types of crimes over regions. Squares depict the pair-wise correlations between crime categories.}
	\label{fig:case_study_att}
\end{figure}

\begin{figure}[h]
	\centering
	\subfigure[][\scriptsize{Hyperedge 7}]{
		\centering
		\includegraphics[width=0.3\columnwidth]{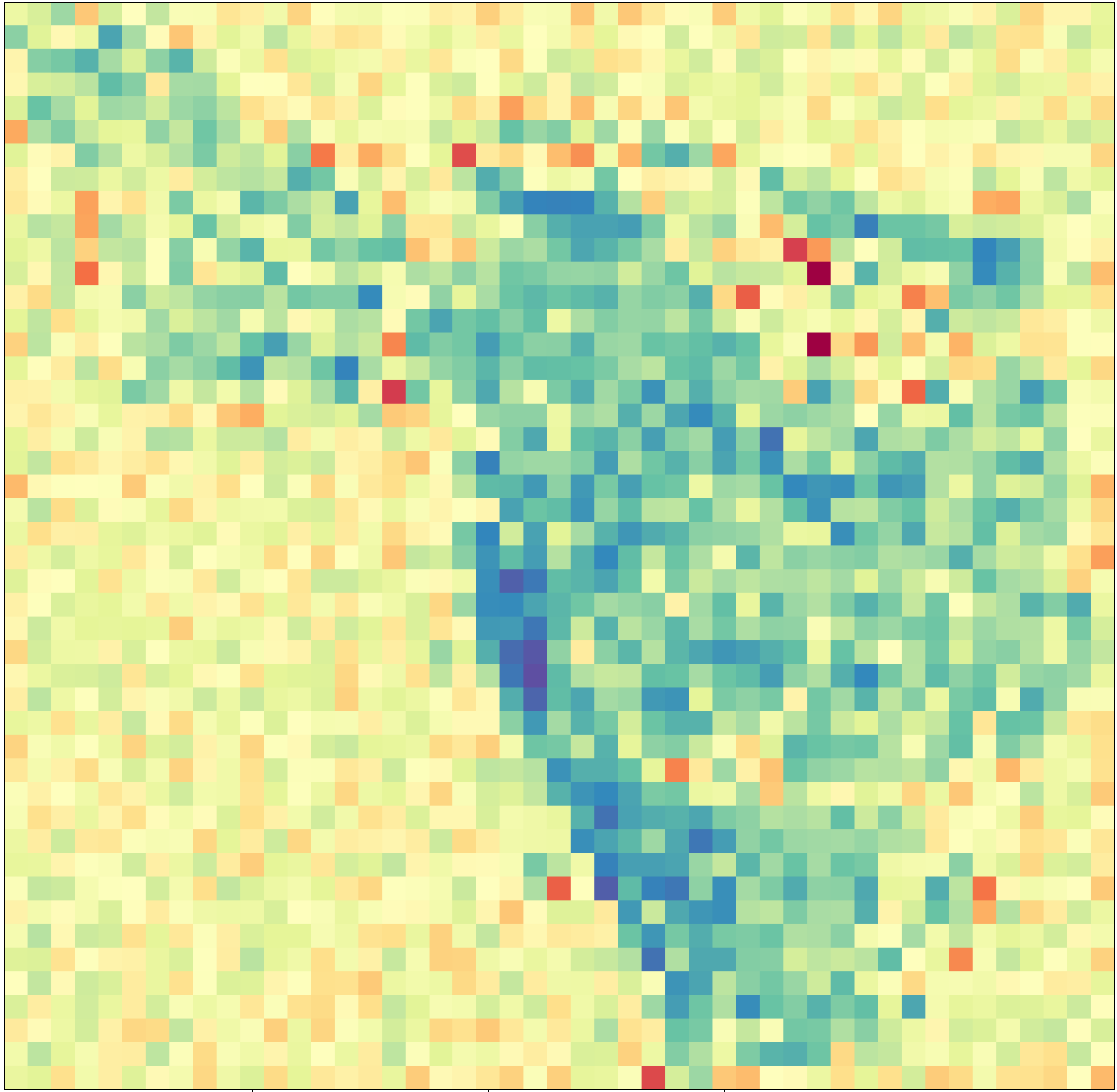}
		\label{fig:case_study_map_mine7}
	}
	\subfigure[][\scriptsize{Hyperedge 8}]{
		\centering
		\includegraphics[width=0.3\columnwidth]{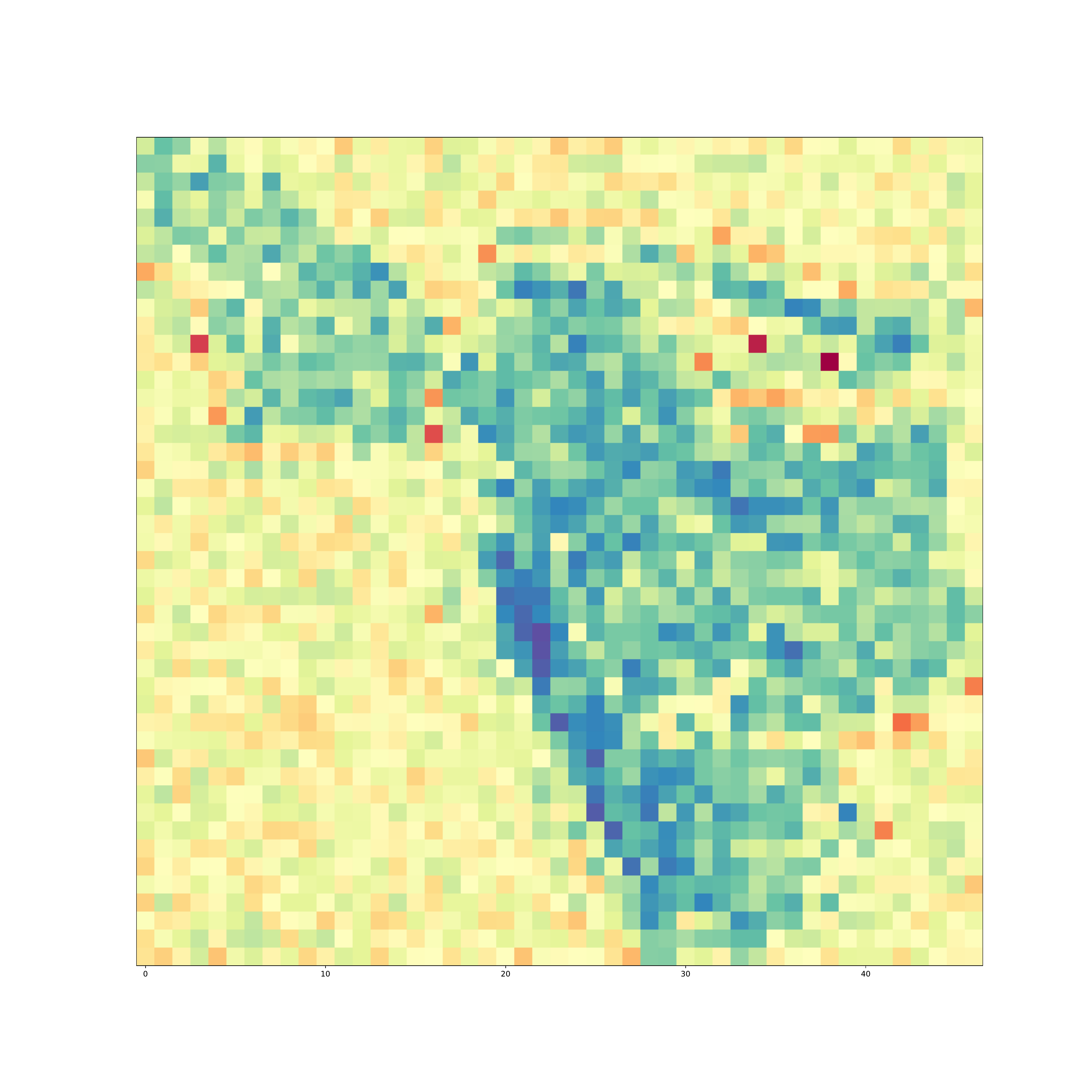}
		\label{fig:case_study_map_mine8}
	}
	\subfigure[][\scriptsize{Crime Distribution}]{
		\centering
		\includegraphics[width=0.3\columnwidth]{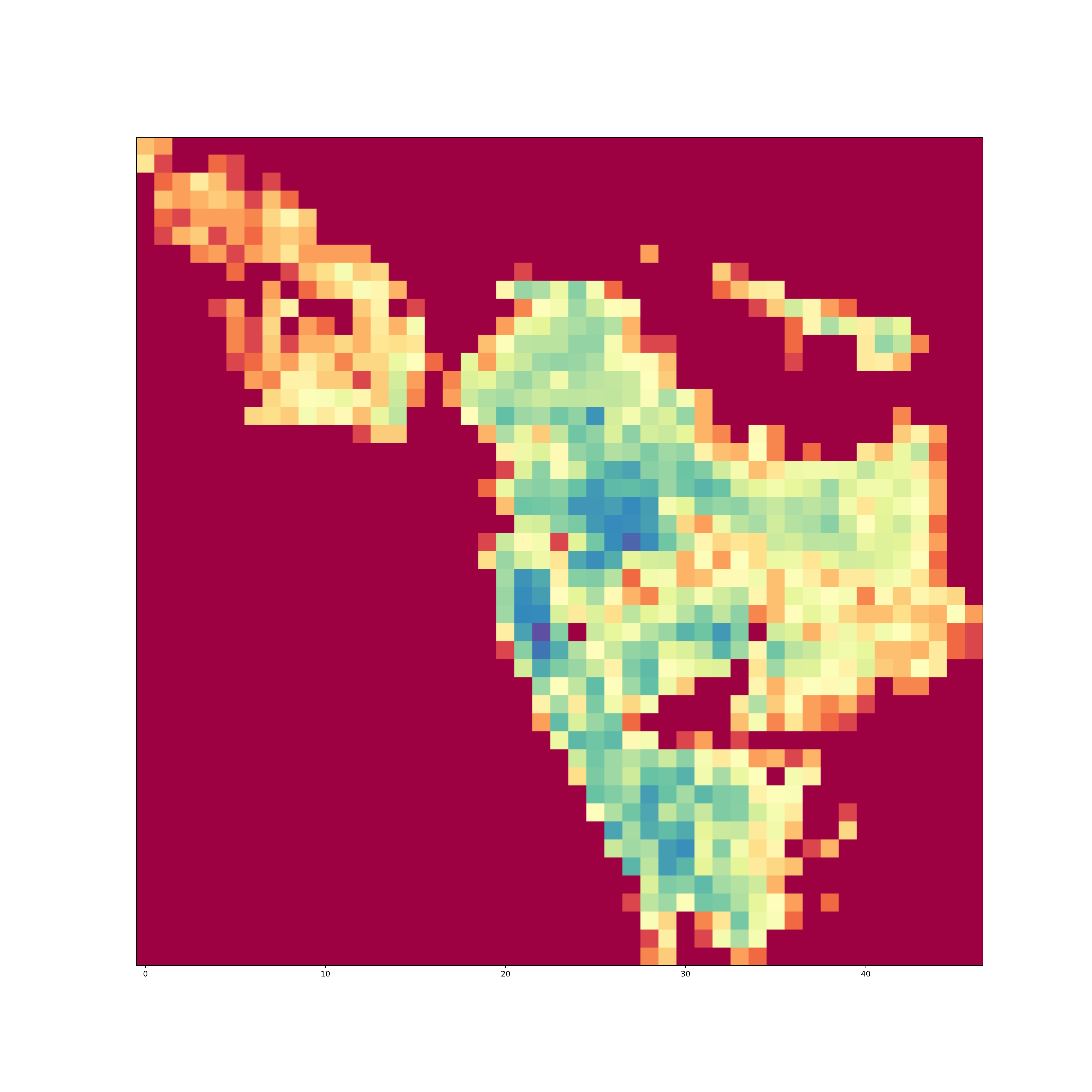}
		\label{fig:case_study_map_benchmark0}
	}
	\caption{Heat maps for the encoded relations between all regions and two sampled hyperedges with the joint analysis of actual crime burglary spatial distribution in NYC.}
	\label{fig:case_study_map}
\end{figure}

We offer several examples to give an intuitive impression of our model explainability. Specifically, Figure~\ref{fig:case_study_att} visualizes i) different types of crime distribution (\ie~heat-map-based statistics) from top-3 regions ranked by their learned relevance values with hyperedges (\eg, $e_8$, $e_{41}$); ii) learned type-specific crime influence scores for message propagation in our multi-channel routing mechanism over spatial and temporal dimensions. \emph{First}, we can see that \emph{\model} enables the dynamic modeling of multiplex spatial-temporal dependencies for different types of crimes, which is helpful to provide relation interpretations for policy makers. \emph{Second}, the observed similar distributions among the most relevant regions, indicates that we endow our \emph{\model} with the capability of identifying dependent regions which are either spatially neighbors or not, by capturing local and global geographical relation structures. In addition, we visualize the relevance matrix between a specific hyperedge and all regions in Figure~\ref{fig:case_study_map_mine7} and Figure~\ref{fig:case_study_map_mine8}. By jointly analyzing the visualized relevance matrix and the actual crime spatial distributions across different regions, we can notice that the consistent distribution patterns suggest that \emph{\model} could well differentiate regions with their underlying crime patterns.


\section{Related Work}
\label{sec:relate}

\paragraph{Spatial-Temporal Prediction.}Many prediction methods have been proposed to uncover the implicit dependencies between future data points and historical observations from various spatial-temporal data~\cite{li2020autost,2019dynamicfusion}. Specifically, attention mechanism has been used to enhance the temporal representation of recurrent neural network~\cite{yao2019revisiting,feng2018deepmove,2020cross}. Furthermore, convolution-based sequential models propose to capture the spatial-temporal information simultaneously~\cite{fengsequential,zhang2017deep}. In this work, we aim to predict crimes by exploring inherent dependencies across time, location and semantic dimensions. 

\paragraph{Graph Neural Networks for Spatial-Temporal Data.}The objective of graph neural networks is to project nodes from the generated graph into a low-dimensional vector space~\cite{wang2019heterogeneous,hu2020heterogeneous}. It is worth mentioning that several recent efforts provide insights into graph neural networks for mining spatial-temporal data~\cite{song2020spatial,diao2019dynamic}. For example, ST-GCN~\cite{yu2017spatio} takes inspiration from graph convolution network to model inter-region spatial dependency over the generated geography graph. DCRNN~\cite{li2018diffusion} captures the spatial correlations by performing the bidirectional random walks on the region graph. However, these methods fall short in encoding region-wise relations in a dynamic manner, due to their utilization of static spatially adjacency matrices.


\section{Conclusion}
\label{sec:conclusion}

In this paper, we propose a novel \full\ (\model), which approaches better crime pattern learning from spatial, temporal and semantic dimensions. \model\ is equipped with multi-channel routing mechanism to capture region-wise relations at a fine-grained level and exhibit the explainable semantics. Furthermore, with the relational path-aware message aggregation, \model\ refines the spatial-temporal representations under the context of long-range and global dependencies. Experiments demonstrate that the proposed method significantly outperforms various baselines.
In the future, we would like to extend our model for real-time crime prediction with streaming data.

\section*{Acknowledgments}
We thank the anonymous reviewers for their constructive feedback and comments. This work is supported by National Nature Science Foundation of China (62072188), Major Project of National Social Science Foundation of China (18ZDA062), Science and Technology Program of Guangdong Province (2019A050510010).

\clearpage
\bibliographystyle{named}
\bibliography{ijcai21}

\end{document}